# SentiMaithili: A Benchmark Dataset for Sentiment and Reason Generation for the Low-Resource Maithili Language

RAHUL RANJAN*, MAHENDRA KUMAR GURVE*, ANUJ, and YAMUNA PRASAD, Indian Institute of Technology Jammu, India

NITIN, University of Cincinnati, USA

Developing benchmark datasets for low-resource languages poses significant challenges, primarily due to the limited availability of native linguistic experts and the substantial time and cost involved in annotation. Given these challenges, Maithili is still underrepresented in natural language processing research. It is an Indo-Aryan language spoken by more than 13 million people in the Purvanchal region of India, valued for its rich linguistic structure and cultural significance. While sentiment analysis has achieved remarkable progress in high-resource languages, resources for low-resource languages, such as Maithili, remain scarce, often restricted to coarse-grained annotations and lacking interpretability mechanisms. To address this limitation, we introduce a novel dataset comprising 3,221 Maithili sentences annotated for sentiment polarity and accompanied by natural language justifications. Moreover, the dataset was carefully curated and validated by linguistic experts to ensure both label reliability and contextual fidelity. Notably, the justifications are written in Maithili, thereby promoting culturally grounded interpretation and enhancing the explainability of sentiment models. Furthermore, extensive experiments using both classical machine learning and state-of-the-art transformer architectures demonstrate the dataset's effectiveness for interpretable sentiment analysis. Ultimately, this work establishes the first benchmark for explainable affective computing in Maithili, thus contributing a valuable resource to the broader advancement of multilingual NLP and explainable AI.

## 1 Introduction

Large Language Models (LLMs) have significantly transformed the landscape of Natural Language Processing (NLP), enabling cutting-edge performance across a wide range of tasks including machine translation, text summarization, sentiment analysis, question answering, and natural language inference. Moreover, LLMs are typically trained with billions of parameters on extensive multilingual corpora. This large-scale pretraining facilitates advanced capabilities in semantic representation, controlled text generation, and context-aware manipulation of natural language. Models

*Both authors contributed equally to this research.

Authors' Contact Information: Rahul Ranjan, rahul.ranjan72@hotmail.com; Mahendra Kumar Gurve, mahendra.gurve@iitjammu.ac.in; Anuj, Anuj@iitjammu.ac.in; Yamuna Prasad, yamuna.prasad@iitjammu.ac.in, Indian Institute of Technology Jammu, Jammu, J&K, India; Nitin, University of Cincinnati, Cincinnati, USA, nitinfu@ucmail.uc.edu.

such as GPT[39], BERT [12], and T5 [32] have thus positioned LLMs as the backbone of modern NLP systems. However, the success of these models has been predominantly concentrated in high-resource languages such as English, Chinese, French, and Spanish. These languages are characterized by abundant digital content, rich annotated datasets, extensive user communities, and strong institutional support. The availability of such resources enables effective pretraining and fine-tuning of models, thereby accelerating progress in the development of language-specific tools and applications.

On the other hand, while high-resource languages continue to benefit from rapid advancements in NLP, low-resource languages face significant barriers to inclusion. These languages often lack the essential tools required for computational tasks, such as sentiment analysis . Annotated datasets, pretrained language models, and standardized linguistic resources are either limited or entirely absent. Additionally, many low-resource language communities are becoming increasingly active online, particularly on platforms like Twitter and Facebook, where they engage in social and political discourse. Despite this growing digital presence, such languages remain underrepresented in the NLP research landscape. Many languages are spoken in developing regions, including parts of the Indian subcontinent, where understanding public sentiment is crucial for businesses, governments, and civil society. However, the scarcity of digitized corpora, labeled data, and orthographic standards continues to hinder model development. This technological gap not only restricts research progress but also excludes millions of speakers from AI-enabled applications such as information access, content moderation, and digital assistance.

Moreover, addressing the challenges faced by low-resource languages is not only a technical necessity but also an ethical imperative. The AI and NLP communities must ensure linguistic inclusivity, especially as language technologies become central to education, governance, healthcare, and communication. Developing tools for these languages promotes digital literacy, preserves linguistic diversity, and empowers communities to engage in the digital world. Additionally, the unique linguistic and cultural features of low-resource languages enrich our understanding of human language and improve the adaptability of multilingual models.

Among the many low-resource languages facing critical gaps in NLP research, Maithili emerges as a linguistically significant yet computationally underserved language. It is an Indo-Aryan language with a substantial speaker base and rich cultural heritage. According to the Census of India, approximately 13.5 million people speak Maithili, primarily in the state of Bihar and adjacent regions of Nepal. Although Maithili is officially recognized as one of the 22 scheduled languages of India, it remains critically underrepresented in the natural language processing ecosystem [37]. Specifically, it lacks coverage in standard multilingual benchmarks, and there is a severe scarcity of annotated corpora, pretrained models, and linguistic tools. As a result, foundational NLP tasks such as sentiment analysis, machine translation, and syntactic parsing remain difficult to implement effectively. The absence of these resources not only limits the development of language technologies but also restricts the digital inclusion of Maithili speaking communities. Therefore, addressing the computational gaps in Maithili is essential for promoting equitable access to AI-driven applications and advancing the broader goals of multilingual NLP.

To bridge the gap in low-resource language processing, this paper presents a new benchmark dataset for Maithili, comprising sentence-level sentiment labels and corresponding justification annotations curated by linguistic experts. To mitigate the impact of data scarcity, the proposed approach leverages pretrained models adapted to low-resource settings. Building on this resource, a sentiment classification model is developed using a large language model (LLM) architecture specifically designed for low-resource scenarios. Furthermore, a Hierarchical Reasoning Architecture is introduced, which operates in two stages: the first stage predicts the sentiment label, and the second stage generates a justification conditioned on both the input sentence and the predicted sentiment. This architecture facilitates interpretable sentiment analysis by coupling classification with rationale generation. The proposed framework aims not only to advance NLP capabilities for Maithili but also to serve as a scalable solution for other low-resource languages.

The key contributions of this research are as follows:

- **Construction of a Benchmark Dataset:** A high-quality annotated dataset for Maithili is introduced, consisting of sentence-level sentiment labels and corresponding justification texts curated by linguistic experts. This resource addresses the lack of supervised data for low-resource language modeling.
- **Development of a Low-Resource Sentiment Classifier:** The sentiment classification model is implemented using a large language model (LLM) architecture optimized for low-resource settings, leveraging multilingual pretraining and transfer learning techniques.
- **Two-Stage Hierarchical Reasoning Architecture:** A two-stage architecture is proposed that first performs sentiment classification and subsequently generates textual justifications conditioned on the input and predicted label, supporting interpretable and explainable NLP.
- **Comprehensive Experimental Validation:** Rigorous experiments are conducted to evaluate the performance of the proposed models on sentiment classification and justification generation tasks. The results are benchmarked against multiple baselines to demonstrate effectiveness.

The structure of this paper is as follows: Section 2 reviews related work on sentiment analysis and justification generation in Maithili and other low-resource language settings. Section 3 describes the construction and annotation process of the Maithili dataset. Section 4 outlines the proposed methodology, including the two-stage sentiment and justification framework. Section 5 and 6 present the experimental setup and results, respectively . Section 7 concludes the paper with key findings and future directions.

## 2 Related Work

This section reviews relevant literature across four key areas: sentiment analysis in low-resource languages, approaches in Indian languages, available resources for Maithili, and models that incorporate justification generation for explainable sentiment analysis.

### 2.1 Sentiment Analysis in Low-Resource Languages

Sentiment analysis in low-resource languages faces persistent challenges due to the absence of large-scale annotated corpora, linguistic tools, and pretrained models. Moreover, language models such as mBERT [12] and XLM-R [8] attempt to bridge this gap through cross-lingual transfer learning. However, their performance decreases dramatically when applied to languages not well represented in their pretraining data. Recent studies suggest that relying on language-specific resources remains essential for reliable NLP outcomes. For instance, Koto et al. [21] demonstrate that incorporating multilingual sentiment lexicons during pretraining can significantly improve zero-shot sentiment classification across 34 languages, including several low-resource and code-mixed languages. Similarly, Aliyu et al. [30] benchmark multiple transformer models (e.g., AfriBERTa) on 12 African languages, showing the superiority of transformer-based architectures in low-resource contexts.

### 2.2 Sentiment Analysis in Indian Languages

A range of sentiment resources now exist for Indian languages. For instance Akhtar et al. [2] compiled an 8,000-sentence corpus from news, blogs, and e-commerce reviews for the Bangali language sentiment analysis, Kumar et al. [23] introduced the 20,000-sentence BHAAV dataset from short Hindi stories.

Gangula and Mamidi[15] introduced Sentiraama for Telugu, a dataset consisting of 1,000 documents drawn from diverse domains. In Bengali, Karim and Cochez [19] curated 320,000 annotated documents, followed by the 15,000-instance SentNoB corpus [16]. More recently, IndicSentiment [13] expanded sentiment resources to 13 Indian languages. Lexical resources like the Hindi Subjective Lexicon [4] and systems such as IIT-TUDA [22] advanced sentiment classification. Akhtar et al. [1] explored Aspect-Based Sentiment Analysis (ABSA) for aspect category detection and sentiment classification for Hindi.

### 2.3 Maithili Language Resources

Low-resource Indian languages such as Bhojpuri, Maithili, and Magahi have recently seen the creation of annotated corpora for POS and chunking, enabling basic NLP tool development. Similar efforts for other Indian languages, like Bangla, highlight the importance of structured corpora for linguistic analysis and language technology. [29] Maithili remains significantly under-resourced. Initial NLP efforts by Singh & Jha [17] developed POS-tagged and chunked corpora. Morphological tools using suffix stripping and finite-state methods were later introduced [33]. POS tagging methods using the Vowel Ending Approach and instance-based learning offer additional tools [31]. Despite these developments, sentiment analysis in Maithili is unexplored. While cross-lingual strategies like Linked WordNets [5]

### 2.4 Sentiment Analysis with Justification

Recent research in sentiment analysis emphasizes *explainability*, where models not only predict sentiment but also generate human-readable justifications for their decisions [3, 7]. These neural architectures, often based on pretrained transformers, are trained on corpora annotated with both sentiment labels and corresponding textual rationales. For example, the e-SNLI dataset enables models to justify entailment decisions, enhancing transparency [7]. Likewise, the SOUL benchmark highlights the justification generation gap between human and model performance, even in advanced systems like GPT-4.

Explainable sentiment analysis has recently expanded to low-resource and morphologically complex languages. Mabokela et al. [28] present a multilingual transformer framework enhanced with LIME and SHAP for sentiment tasks in under-resourced African languages. Similarly, Nazeem et al. [36] demonstrate that interpretable sentiment models for Malayalam, integrated with LIME explanations, significantly improve user trust and system transparency. These studies highlight the growing role of XAI tools in building accountable NLP systems for low-resource settings.

However, Indian languages—particularly Maithili—lack justification-labeled sentiment datasets. Cross-lingual approaches, such as those using linked WordNets for Hindi and Marathi [5], struggle to produce reliable in-language explanations. Their effectiveness further declines in low-resource scenarios due to linguistic and scriptural divergence. Recent work by Rizvi et al. [35] integrates explainable AI into sentiment classification pipelines for Sinhala and code-mixed content, showing promise for multilingual justifications. Nevertheless, such architectures have yet to be explored for Maithili.

Our work addresses this gap by constructing the first manually annotated sentiment-justification dataset for Maithili. It enables the training and evaluation of interpretable sentiment models in a linguistically rich, resource-scarce context, contributing to more inclusive and explainable NLP.

## 3 Dataset Construction and Annotation

This section outlines the methodology for constructing a high-quality sentiment and rationale-annotated dataset in the Maithili language. The process includes multilingual data sourcing, comprehensive preprocessing, expert-driven annotation, data augmentation, and linguistic validation. Given the low-resource status of Maithili, the methodology prioritizes linguistic fidelity, cultural relevance, and annotation transparency to support robust sentiment modeling. Our research in the fields of Deep Learning, AI, and Machine Learning for the Maithili language has been limited by the lack of available datasets. To address this gap, we are introducing a novel dataset.

### 3.1 Multisource Data Acquisition

Maithili is a morphologically rich and syntactically diverse language. However, it suffers from a notable scarcity of publicly available annotated datasets for sentiment analysis—especially those that include explanatory justifications. The inherent linguistic complexity of Maithili further exacerbates these challenges, creating substantial barriers for supervised learning in this domain.

To address these limitations, this paper utilizes a dataset construction process that integrates diverse textual sources featuring natural Maithili usage. Specifically, the dataset incorporates content from regional Wikipedia[1], public discussion forums[2], social media platforms[3], YouTube comments[4], and web libraries[5].

Educational resources such as Maithili-language NCERT materials are also included.[6] [7] Some data used in this study is sourced from the Sentiment140 dataset[8], which contains English-language tweets labeled for sentiment. To align with the Maithili language objectives of this work, a portion of this dataset was translated into Maithili using the IndicTrans2 neural machine translation model. This step allowed for the augmentation of the dataset with synthetically generated sentiment-labeled content in Maithili, helping to further diversify and expand the resource base. This multi-domain strategy effectively captures both formal and informal language registers. As a result, the constructed dataset represents a comprehensive sample of contemporary Maithili language usage.

### 3.2 Preprocessing and Data Curation

Furthermore, all collected texts undergo a multi-stage data preprocessing pipeline. This pipeline aims to enhance linguistic coherence and semantic relevance. Specifically, it involves removing noisy entries, filtering out non-Maithili content, and excluding structurally incomplete or sentiment-neutral utterances. Consequently, the resulting dataset maintains higher quality and consistency for effective sentiment analysis.

---

[1] https://en.wikipedia.org/wiki/Maithili_language
[2] https://khattarkaka.com/kanyadaan; https://www.omniglot.com/language/phrases/maithili.php; https://maithililyrics.wordpress.com/tag/maithili-songs-lyrics/; https://maithilisamachar.com/maithili-stories/page/2/
[3] https://x.com/TeamMithilaRajy; https://x.com/TourismBiharGov; https://x.com/Mithila_Talkies; https://x.com/RajanTirhutiya; https://x.com/chadmaithil
[4] https://www.youtube.com/
[5] https://bloomlibrary.org/language:mai; https://wortschatz.uni-leipzig.de/en/download/Maithili
[6] NCERT Class 10: https://ncert.nic.in/ebsb/10_Maitheli.pdf
[7] SCERT Class 9: https://scert.bihar.gov.in/public/uploads/eresources/CLASS_9_PRAVESHIKA_MAITHLI.pdf
[8] https://www.kaggle.com/datasets/kazanova/sentiment140

These errors negatively affect the fluency and adequacy of the sentences. To ensure high data quality, three linguists with expertise in Maithili are engaged. Among them, two hold doctoral degrees, while one holds a master's degree. The collected data, sourced from diverse platforms, undergoes comprehensive linguistic analysis. Based on this analysis, a set of rules is formulated to systematically correct errors on a sentence-by-sentence basis. Moreover, before implementing the corrections, representative samples from the dataset are provided to the linguists for preliminary evaluation.

Furthermore, these Maithili language experts manually rectify errors in the dataset using a predefined set of linguistic correction rules.

- **Non-Maithili Removal:** Identify and remove entries primarily written in languages other than Maithili.
- **Noise Filtering:** Eliminate sentences containing excessive punctuation, emoticons, special characters, or unintelligible abbreviations.
- **Code-Mixing Regularization:** Standardize mixed-language usage by translating embedded foreign words into equivalent Maithili terms where contextually appropriate.
- **Morphological Correction:** Correct errors related to verb conjugations, tense consistency, noun declensions, and adjective-noun agreement.
- **Structural Completeness:** Discard incomplete sentences or fragments that lack clear semantic intent.
- **Sentiment Clarity:** Exclude ambiguous utterances lacking explicit emotional or evaluative markers.
- **Orthographic Consistency:** Correct spelling inconsistencies, typographical errors, and standardize script usage to Devanagari.
- **Lexical Normalization:** Replace dialect-specific or colloquial expressions with standardized Maithili vocabulary to ensure uniformity.
- **Punctuation Standardization:** Ensure correct and consistent punctuation use to enhance readability and semantic clarity.
- **Semantic Verification:** Confirm that sentence corrections preserve original meaning and context without altering intended sentiment.

Applying these rules systematically ensures improved linguistic coherence and data quality for effective sentiment analysis and justification tasks. Table 1 illustrates common errors in the dataset along with their corresponding corrections made by linguists.

Table 1. Examples of Common Error Types and Their Corrections in Maithili Dataset

| Error Type | Sentence | Correction |
| --- | --- | --- |
| Non-Maithili content | This product bahut achha hai. | ई वस्तु बहुत नीक अछि। |
| Orthographic inconsistency | ओ हमर घरे आबैत अछि. | ओ हमरा घर आबैत अछि। |
| Morphological error (verb conjugation) | हम ओतय गेल छल। | हम ओतय गेल रही। |
| Code-mixing irregularity | ई movie हम पसंद करैत छी। | ई चलचित्र हम पसंद करैत छी। |
| Incomplete structure | ओ बजार सँ। | ओ बजार सँ आबि रहल अछि। |
| Dialect-specific usage | हमरे गाम मे बहुत गरमी अछि। | हमर गाँव मे बहुत गरमी अछि। |
| Noise (special characters, excessive punctuation) | अहाँ कोना छी???!!! | अहाँ कोना छी? |
| Lexical inconsistency | सङ्गीत सुनबाक अछि। | संगीत सुनबाक अछि। |

## 3.3 Annotation Methodology

Furthermore, after the cleaning and preprocessing step described in the previous subsection, where a set of linguistic rules is applied to ensure consistency and quality, each sentence is manually reviewed and corrected by linguistic experts with a sentiment label, either positive or negative. Sentences that express neutral sentiment are excluded to maintain clear polarity distinctions. In addition to sentiment labels, the experts provide a corresponding justification that captures the emotional intent behind each sentence. For example, the sentence "ई सेवा बहुत संतोषजनक अछि।" is annotated as positive, with the justification "'संतोषजनक' शब्द सकारात्मक अनुभव दर्शबैत अछि।"

All datasets undergo rigorous review by domain experts in Maithili sentiment analysis. The review panel includes: a published Maithili-Hindi author with over ten books, a government language teacher with a Ph.D., and an assistant professor holding a Ph.D. The efforts of these linguistic experts are acknowledged in Section 8, along with their detailed profiles. Furthermore, multilingual large language models ChatGPT is used to generate initial justifications. This justification dataset is further verified and corrected by linguistic experts serving as human annotators. This technique reduces manual effort and speeds up annotation. However, nearly all outputs still require revision. Importantly, these corrections ensure cultural relevance, eliminate hallucinations, and maintain consistency with sentiment labels and grammatical rules. This hybrid method balances efficiency with quality. As a result, the final corpus achieves a low error rate.

## 3.4 Annotation Quality Assessment

To assess the reliability of manual annotations in the Maithili dataset, inter-annotator agreement is measured using Cohen's Kappa coefficient. This statistical metric captures agreement beyond chance, offering a more reliable measure than simple percentage match. Each sentence in the dataset is labeled for binary sentiment (positive or negative) by three independent annotators. Subsequently, pairwise Kappa scores are computed across all annotator pairs, as shown in Table 2. A high average Kappa value reflects consistent labeling and supports the validity of the sentiment annotations. Furthermore, to construct the final dataset, majority voting is applied for each sentence. Instances with at least two matching labels are retained. This ensures label consistency and filters out ambiguous or unresolved cases.

Table 2. Inter-Annotator Agreement (Cohen's $\kappa$) on Maithili Sentiment Annotations

| Agreement Between | Cohen's $\kappa$ | Agreement Level |
|---|---|---|
| Expert 1 and Expert 2 | 0.84 | Strong |
| Expert 1 and Expert 3 | 0.82 | Strong |
| Expert 2 and Expert 3 | 0.83 | Strong |
| **Overall Average** | **0.84** | **Strong** |

## 3.5 Data Statistics

The final Maithili sentiment dataset consists of 1,404 positive and 1,496 negative instances in the training set, along with 172 positive and 149 negative instances in the test set—totaling 3,221 annotated samples. We divided the data in traning and test sets. A detailed summary of the dataset distribution is provided in Table 3. These statistics offer insights into the linguistic richness and variability of the content. Table 4 presents example instances from the final dataset, showcasing the original Maithili sentence, its sentiment label, and the corresponding human-written rationale.

Table 3. Summary Statistics of the Maithili Sentiment Dataset

| Statistic | Value |
| --- | --- |
| Total Samples | 3,221 |
| Training Samples | 2,900 |
| Test Samples | 321 |
| Positive Samples (Train/Test) | 1,496 / 172 |
| Negative Samples (Train/Test) | 1,404 / 149 |

Table 4. Example instances from the Maithili sentiment dataset with labels and justifications

| Sentence (Maithili) | Sentiment | Justification |
| --- | --- | --- |
| हमरा बहुत खुशी भेल। | Positive | एहि वाक्यमे 'खुशी' शब्दक प्रयोग भेल अछि जे सकारात्मक भावना व्यक्त करैत अछि। |
| समाजमे बेरोज़गारीक समस्या बहुत गंभीर भेल जा रहल छैक। | Negative | एहि वाक्यमे 'गंभीर' शब्दक प्रयोग भेल अछि जे नकारात्मक भावना व्यक्त करैत अछि। |
| हमर गामक फुलबाड़ीमे बहुत सुंदर फूल फुलाएल अछि। | Positive | एहि वाक्यमे 'सुंदर' शब्दक प्रयोग भेल अछि जे सकारात्मक भावना व्यक्त करैत अछि। |
| हमर आजुक दिनुका समय बड्ड नीकसँ बीतल। | Positive | एहि वाक्यमे 'नीक' शब्दक प्रयोग भेल अछि जे सकारात्मक भावना व्यक्त करैत अछि। |
| आधुनिक समाजमे एकाकीपन बढ़ल जा रहल अछि। | Negative | एहि वाक्यमे 'एकाकीपन' शब्दक प्रयोग भेल अछि जे नकारात्मक भावना व्यक्त करैत अछि। |
| नयनाकेँ खेलमे बहुत रुचि छैक। | Positive | एहि वाक्यमे 'रुचि' शब्दक प्रयोग भेल अछि जे सकारात्मक भावना व्यक्त करैत अछि। |

## 4 Methodology

This study proposes a two-stage hierarchical framework to perform sentiment classification and justification generation for Maithili sentences. In the first stage, the model predicts the sentiment label of a given input sentence. In the second stage, a justification is generated conditioned on both the input and the predicted label. The two stages are executed sequentially, allowing the system to decouple affective inference from explanation generation while maintaining coherence between the predicted sentiment and its rationale.

### 4.1 Problem Formulation

Let the dataset be defined as $\mathcal{D} = \{(x_i, y_i, r_i)\}_{i=1}^{N}$, where each instance consists of a Maithili input sentence $x_i \in \mathcal{X}$, a sentiment label $y_i \in \mathcal{Y} = \{\text{Positive}, \text{Negative}\}$, and a corresponding justification $r_i \in \mathcal{R}$ that explains the sentiment decision in context of the input. The task is structured as a two-step process. First, a classification function maps the input sentence to a predicted sentiment label, denoted as $\hat{y}_i = \mathcal{F}_{\text{cls}}(x_i)$. Then, a conditional generation function produces a justification based on both the input sentence and the predicted label: $\hat{r}_i = \mathcal{F}_{\text{gen}}(x_i, \hat{y}_i)$. The final output for each instance is a tuple $(\hat{y}_i, \hat{r}_i)$, representing the predicted sentiment and its corresponding justification.

### 4.2 Method Overview

This work presents a two-stage hierarchical framework for sentiment classification and justification generation in Maithili, as shown in Figure 2. The architecture comprises two sequential modules: a sentence-level sentiment classifier and a justification generator. Both modules are built upon pretrained transformer models optimized for Indian languages—IndicBERT[14] for classification and IndicBART[11] for generation. However, the focus of this work is on advancing sentiment analysis and explanation capabilities for low-resource languages, rather than introducing novel model architectures.

Moreover, IndicBERTv2[14] is employed as the encoder backbone for Maithili sentiment classification due to its cross-lingual generalization capabilities and its state-of-the-art performance on the IndicXTREME benchmark [34? ]. The model comprises 278 million parameters and is pretrained on the IndicCorp v2[14] corpus using a combination of masked language modeling (MLM), translation language modeling (TLM), and sentence alignment (SAM) objectives. Its pretraining across 23 Indic languages, including low-resource languages such as Maithili, enables effective representation learning for morphologically rich and syntactically diverse linguistic inputs.

Subsequently, IndicBART-SS[11] is employed for justification generation owing to its high-capacity sequence-to-sequence architecture tailored for Indic languages, including resource-constrained languages such as Maithili. The model is pretrained on the IndicCorp v2 [14] corpus using a combination of denoising autoencoding (DAE) and causal language modeling (CLM) objectives, enabling it to model the conditional distribution $P(\mathbf{y} \mid \mathbf{x})$, where $\mathbf{x}$ denotes the input text and $\mathbf{y}$ the corresponding output sequence. Unlike generic multilingual models such as mBART[26], IndicBART[11] integrates language-specific subword vocabularies and bidirectional attention mechanisms, which contribute to reduced perplexity $\mathcal{P}$ on Indic language tasks.

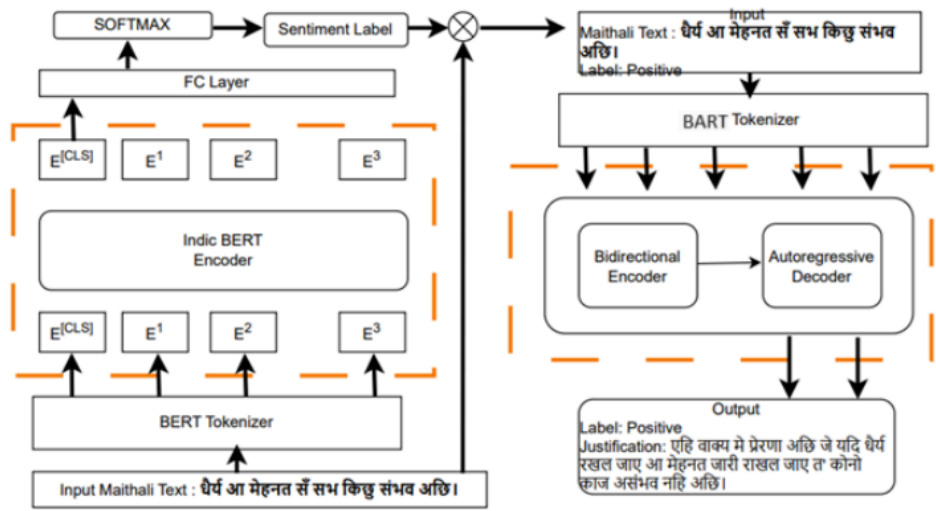

Fig. 1. Two-stage Maithili sentiment analysis pipeline showing: (1) Sentiment classification using IndicBERTv2-SS[14] (left), (2) Justification generation using IndicBART-SSIndicBART[11] (center), with final output (right).

*4.2.1* **Sentiment Classification**. In the first stage, the goal is to predict the sentiment label $\hat{y}_i$ for a given sentence $x_i$. This is achieved using the IndicBERT v2-SS[14] encoder, a bidirectional transformer pretrained on 23 Indian languages using a SentencePiece tokenizer. The sentence $x_i$ is first tokenized as:

$$\mathbf{t}_i = \text{Tok}_{\text{BERT}}(x_i)$$

The tokenized input $\mathbf{t}_i$ is then passed through IndicBERT V2-SS [14] to obtain the contextual embedding $\mathbf{h}_{[\text{CLS}]}$, which represents the entire sentence. A fully connected layer followed by a softmax function maps this embedding to a sentiment distribution:

$$\hat{y}_i = \arg\max\left(\text{Softmax}\left(\mathcal{F}_{\text{cls}}^{\theta}(\mathbf{h}_{[\text{CLS}]})\right)\right)$$

where $\mathcal{F}_{\text{cls}}^{\theta}$ denotes the classification head with parameters $\theta$. Mathematically, given an input sequence $\mathbf{X} = \{x_1, x_2, \ldots, x_n\}$, IndicBERTv2-SS [14] computes contextual embeddings $\mathbf{h}_i = \text{Transformer}_{\theta}(\mathbf{X})_i$ with optimized parameters $\theta$ trained on diverse objectives, enhancing cross-lingual transfer. Notably, IndicBERTv2-SS [14] outperforms monolingual and general multilingual models such as mBERT[12] and XLM-R [9] on sentiment classification tasks, owing to its Indic-specific pretraining. This enables more efficient minimization of the loss function $\mathcal{L}(\mathbf{y}, \hat{\mathbf{y}})$ during fine-tuning. The model's availability on HuggingFace (ai4bharat/IndicBERTv2-SS) ensures reproducibility and ease of deployment.

*4.2.2* **Justification Generation**. In the second stage, the predicted label $\hat{y}_i$ is concatenated with the original input sentence $x_i$ to form a composite input for justification generation:

$$z_i = x_i \oplus \hat{y}_i$$

This combined input $z_i$ is tokenized using the IndicBART-SS[11] tokenizer:

$$\mathbf{z}_i = \text{Tok}_{\text{BART}}(z_i)$$

The tokenized sequence $\mathbf{z}_i$ is then processed by the IndicBART-SS[11] encoder-decoder architecture to generate a justification $\hat{r}_i$:

$$\hat{r}_i = \mathcal{F}_{\text{gen}}^{\phi}(\mathbf{z}_i)$$

where $\mathcal{F}_{\text{gen}}^{\phi}$ is the generation function with parameters $\phi$. The decoder operates autoregressively, predicting each token conditioned on the encoded input and previously generated tokens.

*4.2.3* **Training Setup**. The classification component is trained using a categorical cross-entropy loss:

$$\mathcal{L}_{\text{cls}} = -\sum_{i=1}^{N} \log P(y_i \mid x_i; \theta)$$

The generation module is trained independently using a conditional language modeling objective:

$$\mathcal{L}_{\text{gen}} = -\sum_{i=1}^{N} \sum_{t=1}^{T_i} \log P(r_i^t \mid r_i^{<t}, x_i, \hat{y}_i; \phi)$$

where $T_i$ is the length of the target justification $r_i$, and $r_i^t$ is the token at position $t$.

## 5 Experiment and Results

This section presents the experimental framework and empirical findings of our study. We begin by detailing the experimental setup, including model architectures, training configurations, and evaluation protocols. This is followed by a comparative analysis of state-of-the-art (SOTA) methods for both sentiment classification and justification generation tasks. Additionally, we provide a qualitative assessment of model predictions and generated justifications, with a focus on the Maithili language, highlighting the strengths and limitations of each approach in real-world scenarios.

## 5.1 Experimental Setup

Experiments are conducted using various large language models (LLMs) available through the Hugging Face Transformers library and ML Models. All implementations are developed in Python with PyTorch and executed on a system equipped with CUDA support and an NVIDIA A100 GPU (80GB VRAM), ensuring efficient training and inference for large-scale models.

*5.1.1* **Fine-Tuning and Hyperparameter Optimization:** To ensure optimal model performance, task-specific fine-tuning and systematic hyperparameter tuning are applied to both the classification and generation components.

For IndicBERTv2-SS [14] (sentiment classification), fine-tuning is performed on the Maithili dataset using the AdamW optimizer. A grid search explores learning rates $\{1e^{-5}, 2e^{-5}, 3e^{-5}\}$, batch sizes $\{16, 32, 64\}$, and epochs $\{5, 10, 20\}$.

*5.1.2 Baseline Models.* The following baseline models are used for comparative evaluation:

- **LSTM [10]**: A 2-layer bidirectional LSTM with 256 hidden units in each direction. This serves as a non-transformer baseline for sentiment classification.
- **mBERT [12] (Multilingual Cased)**: A transformer-based model supporting 104 languages, pretrained using masked language modeling and next sentence prediction [12]. It provides a strong multilingual baseline, especially in low-resource scenarios.
- **IndicBERT[14]**: A lightweight ALBERT-based model pretrained on 11 major Indian languages using the IndicCorp corpus [18]. It offers an Indic-centric baseline for classification tasks.
- **mBART[11]**: A multilingual encoder-decoder model developed by Facebook, pretrained with denoising autoencoding for multilingual text generation tasks [25]. It serves as a baseline for justification generation.

*5.1.3 Evaluation Metrics.* The following metrics are used to evaluate the performance of the classification and generation tasks:

- **Classification task**:
  - *Precision*: Measures the proportion of correctly predicted positive instances among all predicted positives.
  - *Recall*: Measures the proportion of correctly predicted positive instances among all actual positives.
  - *F1-score*: Harmonic mean of precision and recall, balancing both aspects. Macro-averaging is used to give equal importance to all classes regardless of frequency.
- **Justification generation task**:
  - *BLEU*: Evaluates precision-based n-gram overlap up to 4-grams between generated and reference text.
  - *ROUGE-1*: Measures unigram (word-level) overlap, emphasizing recall.
  - *ROUGE-L* : Measures the longest common subsequence (LCS) between the generated and reference text.

## 6 Results

This section reports the benchmarking results on the proposed Maithili dataset, evaluated under two experimental settings: zero-shot (without fine-tuning) and task-specific (with fine-tuning) adaptations of multilingual transformer models. A comparative analysis of these settings is provided to assess the impact of fine-tuning on model performance. In addition, the outcomes of the proposed justification pipeline are examined to offer deeper insights into model behavior and overall effectiveness.

**Benchmarking without Model Fine-Tuning**: Table 5 presents the benchmarking results on the proposed Maithili dataset using various multilingual transformer models without fine-tuning. The results clearly demonstrate notable performance variations across the evaluated architectures that support Indian languages. Specifically, MuRIL[20] records the lowest performance, with both accuracy and F1-score at 60.0%, indicating its limited generalization capability for Maithili under zero-shot conditions. In contrast, IndicBERTv1-SS[14] and XLM-R[9] achieve moderately better results of 64.5% and 66.5% respectively, suggesting their comparatively stronger adaptation to low-resource Indian languages. Furthermore, mBERT[12] exhibits a significant improvement with 72.0%, underscoring the effectiveness of its multilingual pretraining in capturing cross-lingual features relevant to Maithili. Extending this progression, the IndicBERTv2[14] variants achieve the highest scores, with the MLM-only version reaching 72.5% and the SS variant further advancing performance to 74.5%, thereby setting a new benchmark for this task. Overall, these findings highlight that advancements in Indic-specific pretraining strategies, particularly those incorporated in IndicBERTv2[14], provide substantial benefits for resource-scarce languages like Maithili when compared to generic multilingual models. Moreover, these results suggest considerable scope for further improvement through fine-tuning on the proposed dataset.

Table 5. Sentiment classification results across multilingual models on Maithili without fine-tuning

| Model | Accuracy (%) | F1-score (%) |
|---|---|---|
| MuRIL [20] | 60.0 ± 0.015 | 60.0 ± 0.015 |
| IndicBERTv1 [14] | 64.5 ± 0.012 | 64.5 ± 0.012 |
| XLM-R [9] | 66.5 ± 0.011 | 66.5 ± 0.011 |
| mBERT [12] | 72.0 ± 0.010 | 72.0 ± 0.010 |
| IndicBERTv2-MLM-only [14] | 72.5 ± 0.009 | 72.5 ± 0.009 |
| IndicBERTv2-SS [14] | 74.5 ± 0.008 | 74.5 ± 0.008 |

**Benchmarking with Model Fine-Tuning:** Table 6 presents the benchmarking results obtained after fine-tuning on the proposed Maithili dataset, which reveal a substantial performance gain across all models compared to the zero-shot setting. Among the baseline methods, LSTM[10] yields the lowest accuracy of 46.0%, reflecting its limited ability to model the complex syntactic and semantic structures of Maithili. Traditional machine learning classifiers such as Naïve Bayes[6] and Random Forest [27] perform moderately better, achieving 62.62% and 64.49% accuracy respectively, but their effectiveness remains constrained due to the absence of contextualized representations. In contrast, fine-tuned transformer models demonstrate significant improvements, with IndicBERTv1[14] and mBERT[12] reaching 83.5% and 85.0%, thereby highlighting the advantages of leveraging contextual embeddings for low-resource languages. XLM-R[9] further improves the results to 92.6%, while MuRIL[20] achieves 94.4%, indicating the positive impact of pretraining on Indian language corpora. The highest performance is observed with the IndicBERTv2[14] variants, where the MLM-only model achieves 95.6% and the SS variant attains 97.2%, establishing a new state-of-the-art benchmark for Maithili. Overall, these results confirm that fine-tuning not only amplifies cross-lingual transfer capabilities but also underscores the effectiveness of Indic-specific pretraining strategies in enhancing performance for resource-scarce languages.

Table 6. Sentiment classification results across models with Fine-Tuning

| Model | Accuracy (%) | F1-score (%) |
|---|---|---|
| LSTM [10] | 46.0 ± 0.021 | 46.7 ± 0.019 |
| Naive Bayes [6] | 62.62 ± 0.008 | 62.61 ± 0.008 |
| Random Forest [27] | 64.49 ± 0.007 | 64.23 ± 0.007 |
| IndicBERTv1 [14] | 83.5 ± 0.011 | 83.3 ± 0.010 |
| mBERT [12] | 85.0 ± 0.008 | 85.0 ± 0.008 |
| XLM-R [9] | 92.6 ± 0.005 | 92.6 ± 0.005 |
| MuRIL [20] | 94.4 ± 0.007 | 94.4 ± 0.006 |
| IndicBERTv2-MLM-only [14] | 95.6 ± 0.006 | 95.6 ± 0.005 |
| IndicBERTv2-SS [14] | **97.2 ± 0.004** | **97.2 ± 0.004** |

**Model Fine-Tuning Impact Analysis:** Figure 2 illustrates a comparative analysis of the performance of all evaluated models with and without fine-tuning on the proposed Maithili dataset. The graph clearly demonstrates the substantial improvements achieved through task-specific adaptation, with all transformer-based models showing significant gains over their zero-shot counterparts. Traditional baselines such as LSTM[10], Naïve Bayes[6], and Random Forest[27] exhibit relatively modest increases, reflecting their limited capacity to leverage contextual information. In contrast, fine-tuned transformers such as IndicBERTv1[14], mBERT[12], and XLM-R[9] achieve marked improvements, while Indian language–specific models like MuRIL[20] and the IndicBERTv2[14] variants demonstrate the highest gains, with the SS variant reaching near-perfect performance. The visualization highlights not only the overall advantage of fine-tuning but also the relative efficacy of Indic-focused pretraining strategies, emphasizing their importance for resource-scarce languages like Maithili.

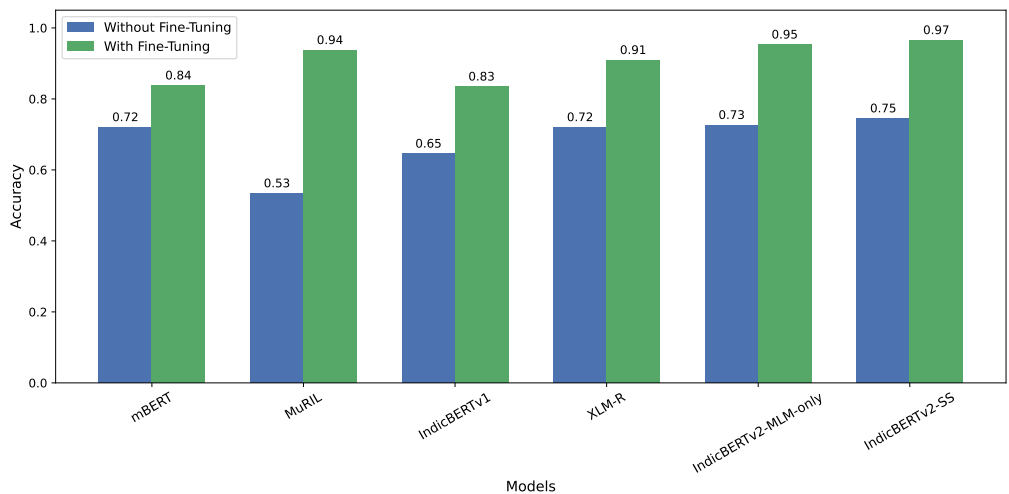

Fig. 2. Accuracy of various models with and without fine-tuning.

**Benchmarking of Justification Generation Models:** Building on its strong performance in the fine-tuned sentiment analysis task, IndicBERTv2 [14] is employed as the primary encoder for the downstream justification generation task. To assess model capabilities, Table 7 presents the evaluation results of several models using BLEU, ROUGE-1, and ROUGE-L metrics. Among the baseline models, BART-MNLI[24] and IndicBART[11] achieve relatively modest performance, with BLEU scores of 17.24 and 20.17, respectively, alongside comparable ROUGE scores, indicating a limited ability to generate accurate justifications. In contrast, models specifically adapted for Indic languages exhibit substantial improvements: IndicBART-SS[11] attains a BLEU score of 34.70, with ROUGE-1 and ROUGE-L scores of 57.77 and 57.27, respectively, while mT5-base[38] achieves a slightly higher BLEU of 35.11 and ROUGE-1/ROUGE-L scores of 56.29 and 55.18. Collectively, these results demonstrate that Indic-specific pretraining, combined with self-supervised adaptation, significantly enhances justification quality, yielding more accurate, contextually relevant, and semantically coherent explanations for sentiment predictions.

Table 7. Justification generation results across models

| Model | BLEU | ROUGE-1 | ROUGE-L |
| --- | --- | --- | --- |
| BART-MNLI [24] | $17.24 \pm 0.010$ | $42.76 \pm 0.010$ | $39.86 \pm 0.009$ |
| IndicBART [11] | $20.17 \pm 0.012$ | $40.74 \pm 0.012$ | $39.19 \pm 0.010$ |
| IndicBART-SS [11] | $34.70 \pm 0.011$ | $\mathbf{57.77 \pm 0.011}$ | $\mathbf{57.27 \pm 0.012}$ |
| mT5-base [38] | $\mathbf{35.11 \pm 0.009}$ | $56.29 \pm 0.009$ | $55.18 \pm 0.010$ |

## 7 Conclusion

This research introduces the first large-scale dataset for interpretable sentiment analysis in Maithili, consisting of expert-annotated sentences with sentiment polarity and native-language justifications. Experimental results with both classical and transformer-based models demonstrate the dataset's effectiveness for improving sentiment analysis in a low-resource setting while also enhancing interpretability. This work establishes a benchmark for affective computing in Maithili and contributes to the advancement of multilingual NLP and explainable AI. In addition, it opens new directions for research in cross-lingual transfer, domain adaptation, and culturally grounded explanation methods.

## 8 Acknowledgment

This work acknowledges the valuable contributions of linguistic experts whose guidance ensured the quality and authenticity of the Maithili annotations. Dr. Narayan Jha, a writer and poet in Maithili and Hindi with over ten published books, also serves as a Maithili language teacher in a Bihar Government High School and holds a Ph.D. from Lalit Narayan Mithila University, Darbhanga. Dr. Radha Kumari, Assistant Professor at Purnea University, Purnia, also contributed her expertise as a Ph.D. holder in Maithili linguistics. Their insights greatly strengthened the cultural and linguistic integrity of the dataset.

**Sentence 1**: जीवन में संघर्षक बाद जे सुख भेटैत अछि, से अनुपम होइत अछि।

**Translation**: The happiness one receives after struggle in life is unique. **Gold Label**: Positive

**Predicted Sentiment**: LSTM: ✗    mBERT: ✗    IndicBERT: ✗    IndicBERTv2: ✓    MuRIL: ✓    XLM-R: ✓    IndicBERT-ss: ✓

**Generated Justifications**:

  BART-MNLI[24]: वाक्य में "अनुपम होइत अछि" शब्द सँ सुख

  mT5-base[38]: एहि वाक्य में जीवन में संघर्षक बाद जे सुख भेटैत अछि, से अनुपम होइत अछ

  IndicBART[11]: जीवन में सकारात्मक भावना व्यक्त हो रहल अछि।

  IndicBART-ss[11]: संघर्षक बाद सुख के अनुपम रूप पर बल देल गेल अछि।

**Sentence 2**: थोड़ा-थोड़ा सन्तोषो जीवन में सच्चा सुख द' सकैत अछि।

**Translation**: Even a little contentment can bring true happiness in life. **Gold Label**: Positive

**Predicted Sentiment**: LSTM: ✗    mBERT: ✓    IndicBERT: ✓    IndicBERTv2: ✓    MuRIL: ✓    XLM-R: ✓    IndicBERT-ss: ✓

**Generated Justifications**:

  BART-MNLI: एहि वाक्यमे सुख शब्दक प्रयोग भेल अछि ज

  mT5-base: वाक्य 'सन्तोषो जीवन में सच्चा सुख द' सकैत अछि' में थोड़-थोड़ सन्तोषक भावना प्रकट भ' रहल अछि।

  IndicBART: थोड़ा-थोड़ा सन्तोषो जीवन में सच्चा सुख द' सकैत अछि जे सकारात्मक भावना के व्यक्त करैत अछि।

  IndicBART-ss: एहि वाक्यमे अनुपम शब्दक प्रयोग भेल अछि जे सकारात्मक भावना व्यक्त करैत अछिन।

**Sentence 3**: जीवन में सुखक नाम पर केवल भ्रम मात्र छै।

**Translation**: In the name of happiness, life is just an illusion. **Gold Label**: Negative

**Predicted Sentiment**: LSTM: ✓    mBERT: ✓    IndicBERT: ✓    IndicBERTv2: ✓    MuRIL: ✓    XLM-R: ✓    IndicBERT-ss: ✓

**Generated Justifications**:

  BART-MNLI: वाक्य में "भ्रम मात्र छै" शब्द सँ असंतो

  mT5-base: एहि वाक्य में जीवन में सुखक नाम पर भ्रम शब्दक प्रयोग भेल अछि जे नकारात्मक भावना व्यक्त करैत अछ

  IndicBART: जीवन में सुखक नाम पर केवल भ्रम मात्र छै आ भ्रम के भावना व्यक्त हो रहल अछि जे नकारात्मक भावना को प्रकट करैत अछि।

  IndicBART-ss: एहि वाक्यमे सच्चा सुख द' सकैत अछि शब्दक प्रयोग भेल अछिए जे सकारात्मक भावना व्यक्त करैत।

**Sentence 4**: सभ किछु हासिल कएला परो, मन में खालीपन रहल — ई की सुख छै?

**Translation**: Despite achieving everything, there was emptiness — is this happiness? **Gold Label**: Negative

**Predicted Sentiment**: LSTM: ✓    mBERT: ✓    IndicBERT: ✗    IndicBERTv2: ✓    MuRIL: ✓    XLM-R: ✗    IndicBERT-ss: ✗

**Generated Justifications**:

  BART-MNLI: वाक्य में "खालीपन रहल" शब्द सँ असंतोष आ

  mT5-base: वाक्य में असंतोष आ निराशा व्यक्त कएल गेल अछि, जतए व्यक्ति किछु हासिल कएला परो, मन में खालीपन रहल — ई की सुख छै?

  IndicBART: सभ किछु हासिल के बावजूद, मन में खालीपन नकारात्मक भावना के संकेत दैत अछि।

  IndicBART-ss: एहि वाक्यमे खालीपन शब्दक प्रयोग भेल अछि जे नकारात्मक भावना व्यक्त करैत अछिपे।

Table 8. Comparison of Sentiment Classification and Explanation Generation across all models after 10 epochs.